\documentclass[a4paper, 10pt, conference]{ieeeconf}      

\IEEEoverridecommandlockouts                              

\usepackage[utf8]{inputenc}

\usepackage{etoolbox}

\usepackage{amsmath}
\usepackage{amssymb}
\usepackage{graphicx}
\usepackage{overpic}
\usepackage{braket}
\usepackage[font=footnotesize]{subcaption}

\usepackage{tikz}
\usepackage{nicefrac}
\usepackage{units}
\usepackage{cite}
\usepackage{framed}

\usepackage{blindtext}
\usepackage[hidelinks]{hyperref}

\usepackage{algorithm2e}
\SetKwSwitch{Switch}{Case}{Other}{switch}{do}{case}{otherwise}{hzn}{end}

\SetCommentSty{mycommfont}
\usepackage{cancel}

\makeatletter
\patchcmd{\algocf@latexcaption}{#3}{#3\endgraf}{}{}
\makeatother

\SetAlCapFnt{\normalfont\footnotesize}
\SetAlCapNameFnt{\normalfont\footnotesize}
\SetAlgorithmName{Alg.}{alg}{Algorithm}

\usepackage{mathtools}

\usepackage{txfonts}
\let\mathbb=\varmathbb

\definecolor{colorFHG}{rgb}{0.09,0.61,0.49}
\definecolor{colorPenalty}{rgb}{0.94,0.27,0.05}
\definecolor{colorPRGreen}{HTML}{97BF20}

\usepackage[textwidth=1.5cm]{todonotes}

\usepackage{aveasmargins}
\IEEEoverridecommandlockouts

\newcommand{\traj}[0]{\xi}

\newcommand{\dee}[0]{\mathrm{d}}

\newcommand{\subs}[1]{_{\mathrm{#1}}}

\newcommand{\transp}{^{\mathsf T}}

\DeclareMathOperator*{\argmin}{arg\,min}

\title{\LARGE \bf
Realtime Global Optimization of a Fail-Safe Emergency Stop Maneuver\\
for Arbitrary Electrical / Electronical Failures in Automated Driving*
}

\author{F. Duerr$^{1}$,  J. Ziehn$^{2,\dagger}$, R. Kohlhaas$^{3}$, M. Roschani$^{2}$, M. Ruf$^{2}$ and J. Beyerer$^{2,4}$\vspace{-10pt}
\thanks{*This publication was written in the framework of KAMO: Karlsruhe Mobility / Profilregion Mobilit\"atssysteme Karlsruhe (\href{https://www.kamo.one}{kamo.one}), which is funded by the Ministry of Science,
Research and the Arts and the Ministry of Economic Affairs, Labour and
Housing in Baden-Württemberg and as a national High Performance Center
by the Fraunhofer-Gesellschaft.}
\thanks{$^{1}$Fabian Duerr is with  Audi AG, 85045 Ingolstadt, Germany
        {\tt\small fabian.duerr@audi.de}}%
\thanks{$^{2}$Juergen Beyerer, Masoud Roschani, Miriam Ruf and Jens Ziehn are with Fraunhofer IOSB, 76131 Karlsruhe, Germany
        {\tt\small \{masoud.roschani, miriam.ruf, jens.ziehn\}@iosb.fraunhofer.de}}%
\thanks{$^{3}$Ralf Kohlhaas is with Robert Bosch GmbH, Corporate Research, Automated Driving,
71272 Renningen, Germany
        {\tt\small ralf.kohlhaas@de.bosch.com}}%
\thanks{$^{4}$Juergen Beyerer is also with the Vision and Fusion Lab, Karlsruhe Institute of Technology KIT, c/o Technologiefabrik, Haid-und-Neu-Strasse~7, 
76131 Karlsruhe, Germany}%
\thanks{$^{\dagger}$Corresponding author}%
}

\begin{document}

\maketitle
\thispagestyle{empty}
\pagestyle{empty}

\aveasSetMargins{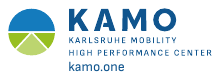}
\aveasSetIEEEFoot{2020}
\aveasSetIEEEHead{%
F. Duerr, J. Ziehn, R. Kohlhaas, M. Roschani, M. Ruf and J. Beyerer, "Realtime Global optimization of a Fail-Safe Emergency Stop Maneuver for Arbitrary Electrical / Electronical Failures in Automated Driving," 2020 IEEE 23rd International Conference on Intelligent Transportation Systems (ITSC), Rhodes, Greece, 2020
}{%
10.1109/ITSC45102.2020.9294578%
}

\begin{abstract}
In the event of a critical system failures in automated vehicles, fail-operational or fail-safe measures provide minimum guarantees for the vehicle's performance, depending on which of its subsystems remain operational. Various such methods have been proposed which, upon failure, use different remaining sets of operational subsystems to execute maneuvers that bring the vehicle into a safe state under different environmental conditions. One particular such method proposes a fail-safe emergency stop system that requires \emph{no} particular electric or electronic subsystem to be available after failure, and still provides a basic situation-dependent emergency stop maneuver. This is achieved by preemptively setting parameters to a hydraulic / mechanical system prior to failure, which after failure executes the preset maneuver ``blindly''. The focus of this paper is the particular challenge of implementing a lightweight planning algorithm that can cope with the complex uncertainties of the given task while still providing a globally-optimal solution at regular intervals, based on the perceived and predicted environment of the automated vehicle.
\end{abstract}

\section{INTRODUCTION}\label{sec:introduction}

The optimization of emergency maneuvers has been the subject of comprehensive research, with a wide range of solutions addressing different notions of what constitutes an ``emergency'' in this context: From a fully operational vehicle encountering challenging environmental conditions (e.g. \cite{werling2012automatic} for model-predictive pedestrian avoidance; \cite{magdici2016fail} for reactions to unexpected traffic situations; \cite{frese2011comparison} for collision avoidance for cooperative vehicles), to various stages of degraded capabilities of subsystems under various internal and external conditions (e.g. \cite{reschka2016safety, itsc}), including incapacitation of the responsible human driver for systems up to SAE level 3 (e.g. \cite{takahashi2016automated, becker2014bosch}).

Degraded capabilities of the ego vehicle include single- or multiple-point faults, which can be addressed by redundant systems to provide a fail-operational behavior (cf. \cite{Ishigooka}), possibly including graceful degradation (cf. \cite{kim2013towards}) by reducing active vehicle functions depending on the occurring failure modes, or at least fail-safe behavior, which provides minimal functions to assure safety in case of a failure. In each case, the chosen fallback behavior depends on the assumed set of remaining operational systems; a single failed sensor is more easily compensated than a fusion or planning unit; approaches to address various kinds of failure modes are given in \cite{reschka2012, reschka2016safety}.

\begin{figure}%
\vspace{-10pt}
\begin{overpic}{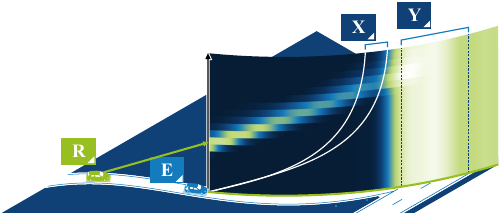}%
\put(45,25){\footnotesize\color{white}$W(t,s)$}
\put(39,31){\footnotesize$t$}
\put(97,11){\footnotesize$s$}
\end{overpic}
\caption{Motivating example: The ego vehicle \textbf{E} is equipped with a hydraulic piston accumulator, whose pressure is released onto the brakes in case of a severe system failure. To provide a situation-dependent maneuver, the pressure is controlled by a valve, which is adjusted at periodical intervals $\Delta t\subs{plan}$ prior to failure, to prepare for a possible emergency. Since failure can occur any time within the upcoming $\Delta t\subs{plan}$ (or never at all), not one single braking trajectory can be planned, but instead a continuous range \textbf{X} of trajectories (and stopping distances) can occur, displaced over failure time. In the given scenario, the safest decision would be to decelerate gently enough to avoid a rear-end collision with car \textbf{R}, yet strongly enough to not enter the road ahead \textbf{Y}. The goal of the proposed algorithm is to minimize the risks $W(t,s)$ over time $t$ and arc length $s$ within the region \textbf{X} with very limited computational effort.}%
\label{fig:motivation}%
\end{figure}

An approach to establish a lower bound of possible safety is proposed in \cite{tuevnotbrems}, where a situation-adaptive emergency stopping maneuver is provided without requiring the use of any electric or electronic system after the moment of failure; the system can therefore be used as a fallback for any failure mode where no superior dedicated solution can efficiently be implemented. To achieve this behavior, motivated in Fig.~\ref{fig:motivation}, it uses a hydraulic~/ mechanical subsystem to brake the vehicle to a halt, and an electronic system, required only prior to failure, which periodically adjusts the hydraulic~/ mechanical system to an optimal, situation-dependent braking deceleration, preemptively for the case of a failure before the next optimization interval.

This paper focuses on the planning task of the described system, addressing the choice of an appropriate planning model, and especially its efficient computational solution, since the purpose of the system as a last-resort fallback demands a lightweight implementation. To this end, Sec.~\ref{sec:system-overview} provides a brief overview of the system presented in \cite{tuevnotbrems}; Sec.~\ref{sec:mathematical-model} establishes the basic mathematical model for the emergency maneuver planning; Sec.~\ref{sec:simplification} describes the proposed approach to render the problem tractable for realtime computation; and Sec.~\ref{sec:problem-solution} describes an efficient approach to solving this problem algorithmically. The performance of the resulting algorithm, which provides a globally-optimal decision for the current planning step, is discussed in Sec.~\ref{sec:practical-results} both in terms of result quality and of computational efficiency. Section~\ref{sec:conclusion} summarizes the main conclusions and provides an outlook to possible future extensions.


\section{SYSTEM OVERVIEW}\label{sec:system-overview}

\begin{figure}[t!]%
\begin{subfigure}{\columnwidth}
\includegraphics{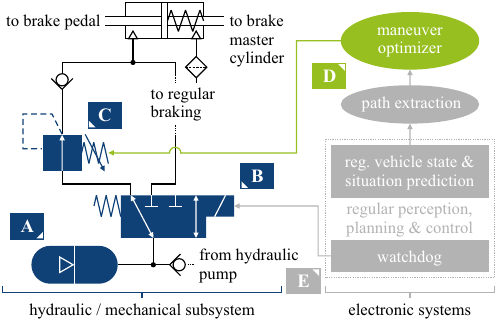}%
\vspace{-2pt}
\caption{Reduced layout of the system originally presented in \cite{tuevnotbrems}. Upon failure of the E/E systems (right), detected by a watchdog mechanism \textbf{E}, the hydraulic / mechanical subsystem (left, originally in \cite{weissenbach}) engages. Valve \textbf{B} opens and releases the pressure from piston accumulator \textbf{A} towards the pressure regulation valve \textbf{C}, whose state is adjusted by the maneuver optimizing unit \textbf{D} (the subject of this paper) at regular intervals to choose the optimal deceleration profile for the vehicle's current situation.}%
\label{fig:hydraulic-electronic}%
\end{subfigure}\\[6pt]

\begin{subfigure}{\columnwidth}
\begin{overpic}[
]{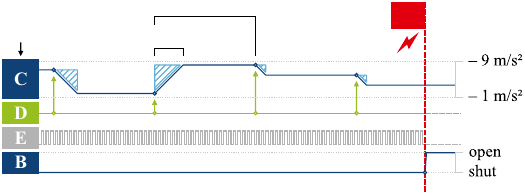}%
\put(0,31.5){\begin{minipage}{2cm}\footnotesize\raggedright part labels \linebreak as in (\subref{fig:hydraulic-electronic})\end{minipage}}
\put(28,29){\footnotesize$\Delta t\subs{valve}$}
\put(35,35.5){\footnotesize $\Delta t\subs{plan}$}
\put(75.7,33.5){\footnotesize \color{white}$t\subs{fail}$}

\put(11,16.5){\footnotesize $a\subs{next}$}
\put(30.5,16.5){\footnotesize $a\subs{next}$}
\put(49.5,16.5){\footnotesize $a\subs{next}$}
\put(69,16.5){\footnotesize $a\subs{next}$}
\end{overpic}\\[-15pt]
\caption{Exemplary timing diagram of the developed system, as in \cite{tuevnotbrems}. Prior to failure, the emergency planning system \textbf{D} computes a new target deceleration $a\subs{next}$ at regular intervals (spaced by $\Delta t\subs{plan}$), used to set a pressure regulation valve \textbf{C}. The valve transitions for some time $t\subs{valve}$ (hatched areas) before reaching $a\subs{next}$. (For simplicty, we consistently denote the hydraulic valve state directly by its associated, calibrated deceleration, in the sense that the valve is preset to achieve this deceleration.) When the watchdog signal ceases, the lock valve \textbf{B} releases the pressure onto valve \textbf{C}, whose current state freezes upon failure and effects a constant braking deceleration.}%
\label{fig:timing}%
\end{subfigure}\\

\begin{subfigure}{\columnwidth}
\includegraphics[width=\columnwidth]{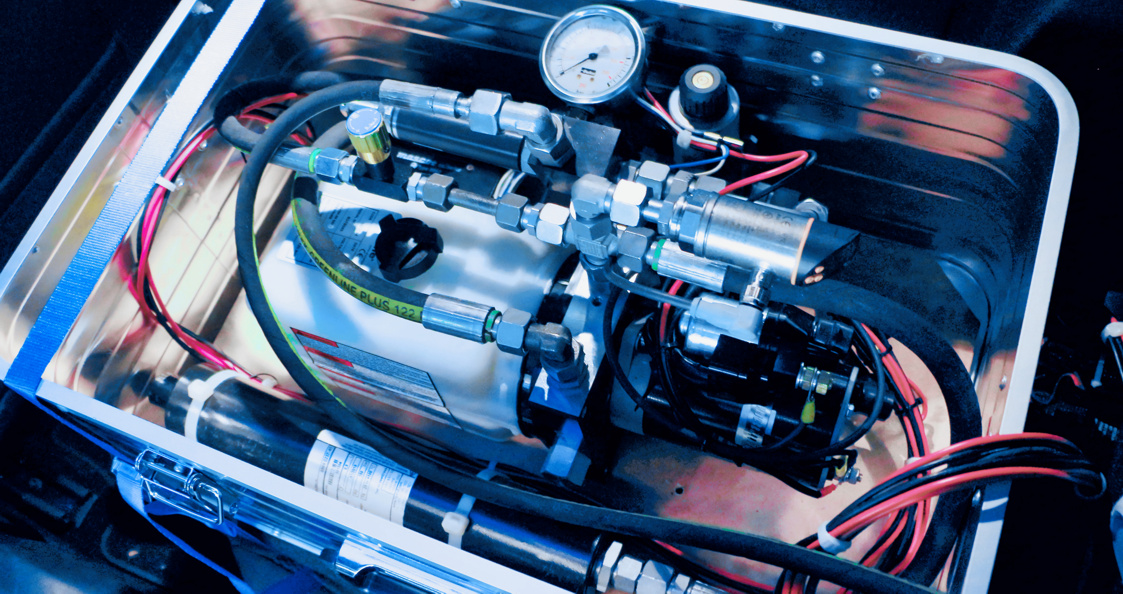}
\caption{Hydraulic prototype of the system in a VW Golf VII Variant, developed by the Institute of Vehicle System Technology (FAST) at KIT.}
\end{subfigure}

\caption{Overview of the system controlled by the optimization algorithm.}
\label{fig:system-overview}
\end{figure}
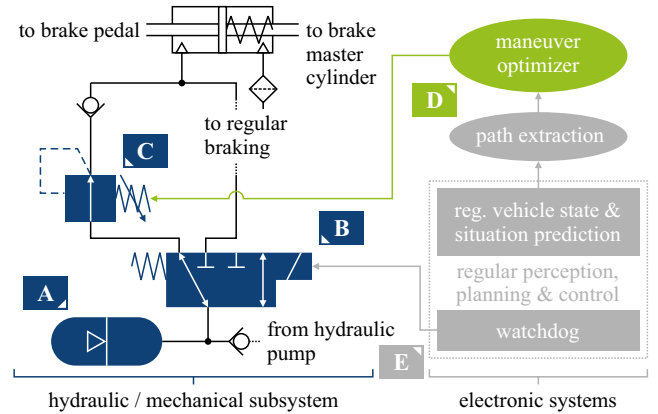
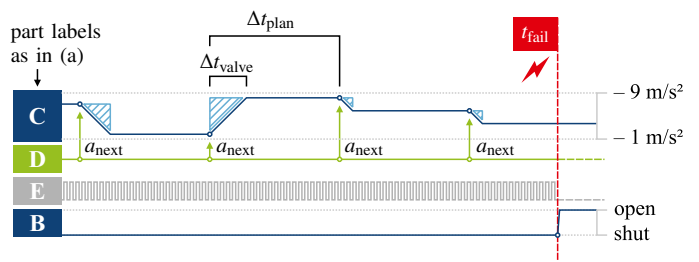
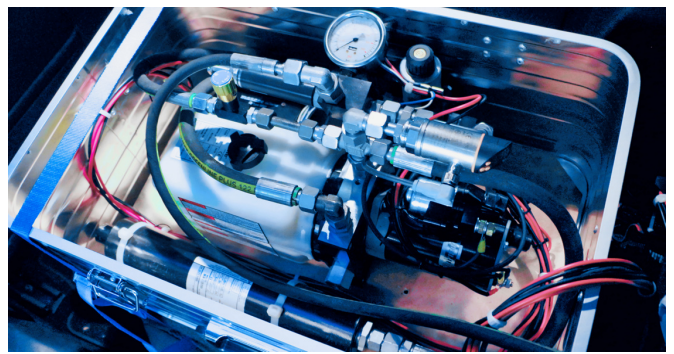

The system's goal is to assure that an ``optimal'', situation-dependent emergency stop maneuver is executed without requiring any electric or electronic components after the time of failure, $t\subs{fail}$. This is achieved, as described in \cite{tuevnotbrems} and shown in Figs.~\ref{fig:system-overview} and \ref{fig:timing}, by electronically presetting hydraulic / mechanical components \emph{prior} to failure in an ``optimal'', situation-dependent way, such that \emph{upon} failure, only hydraulic / mechanical processes are required to execute the preset maneuver. If no failure occurs, the hydraulic / mechanical components remain inactive, and the planned emergency maneuver is not executed.

This section will briefly outline the system, as far as relevant to the maneuver optimizer (\textbf{D} in Figs.~\ref{fig:hydraulic-electronic} and \ref{fig:timing}), whose optimization algorithm is the subject of this paper and will be detailed in Secs.~\ref{sec:mathematical-model} through \ref{sec:problem-solution}.

The optimization algorithm \textbf{D} determines, at regular intervals $\Delta t\subs{plan}$, a target braking deceleration $a\subs{next}$, which is used to adjust a pressure regulation valve \textbf{B}. If the system fails within the current interval (i.e. before the next step of the optimizer), pressure is released immediately, regulated by the valve, to act on the brake master cylinder, executing an emergency stop using the preset $a\subs{next}$.
The optimizer has to determine some $a\subs{next}$ based on the vehicle's current situation at $t\subs{now}$ (e.g. traffic, vehicle dynamics, predictions with uncertainties), conditioned on the assumption that the vehicle fails before the next planning cycle at $t\subs{now}+\Delta t\subs{plan} =: t\subs{plan}$. Since the electric/electronic (E/E) system is still live upon optimization, E/E components can be used to determine $a\subs{next}$, such as processors and data from the vehicle sensors.

On the other hand, the computation must be extremely lightweight, since under typical conditions, the system should rarely ever be required at all; and it must be able to cope with additional complications, arising from uncertainty of the exact time of failure, and from the non-negligible time the pressure regulation valve \textbf{B} takes to reach the state $a\subs{next}$. 

The final requirement is that the optimizer be consistent with a given regular maneuver planner; this allows to naturally specify its key parameters based on the parameters of the regular maneuver planner, and, more importantly, to reuse results to reduce computational effort.

To assure predictability of the \emph{lateral} motion, moderate force is applied to maintain the current steering wheel angle upon failure (while still allowing a human driver to intervene), such that the lateral motion of the vehicle can be assumed to be an arc with known curvature. The optimizer can thus make use of path-velocity decomposition (PVD, \cite{pvd}), without optimizing lateral motion, and accounting only for a limited added degree of positional uncertainty, which further includes uncertainties in perception and prediction, road friction and initial speed due to measurement uncertainties or accelerations at the moment of failure. An experimental evaluation of the predictability of vehicle motion for this use case is provided in \cite{tuevnotbrems}, which describes in more detail how uncertainties are included in the planning process.

\section{MATHEMATICAL MODEL}\label{sec:mathematical-model}

As previously stated, we aim to define the emergency planning problem such that it is consistent with regular maneuver planning of the automated vehicle. Therefore, we first describe the assumed problem statement of the regular maneuver planning, and then derive the statement of emergency maneuver planning from a special case thereof.

\subsection{Regular Planning}

The regular trajectory planning we assume to be modeled as a variational problem, as used e.g. in \cite{ziegler, RZ.2014a, Ruf2018_1000085281}.  In this we consider the trajectory of the ego vehicle to be sufficiently determined by its \emph{trajectory}
\begin{equation}
\traj : [t\subs{now}, t\subs{hzn}] \rightarrow \mathbb R \times \mathbb R, \quad \traj(t) = \left[\traj\subs{x}(t),\; \traj\subs{y}(t)\right]\transp,
\end{equation}
describing the ground coordinates of its rear axle center up to the prediction horizon. With the assumption of negligible tire slip, which strictly aligns the vehicle body with the trajectory's tangent, most common parameters such as heading, yaw rate or individual wheel speeds and angles can be derived given the basic vehicle geometry\cite{icmc}.

As stated in Sec.~\ref{sec:system-overview}, the maneuver is executed with a constant steering wheel angle, to allow for path-velocity decomposition (PVD): We may consider a parametrization of $\traj$ by arc length, its \emph{path} $\bar{\traj}(s)$, together with an appropriate \emph{timing} along this path $\sigma(t)$, such that $\traj(t) \equiv \bar{\traj}(\sigma(t))$. 
In the context of PVD, we assign \emph{penalty costs} to a timing by using a functional of the form
\begin{equation}
\mathcal P[\sigma(\,\cdot\,)] = \int_{t\subs{now}}^{t\subs{hzn}}\!\dee t \; L(t, \sigma(t), \tfrac{\dee}{\dee t}\sigma(t), \tfrac{\dee^2}{(\dee t)^2} \sigma(t), ...),
\label{eq:euler-lagrange}
\end{equation}
where we write $\sigma(\,\cdot\,)$ to denote that the penalty is accumulated over the single parameter $t$ of $s$. For regular trajectory planning, $L$ uses the local (at $t$) derivatives to assign penalty costs e.g. for risks, comfort, traffic rule compliance, efficiency and ecology. For the evaluation of emergency stop maneuvers, we simplify the problem by using $L(t, \sigma(t), \tfrac{\dee}{\dee t}\sigma(t), ...) =: W(t, \sigma(t))$, which is sufficiently expressive to assign \emph{risk penalties} to time--space coordinates that the vehicle should not traverse (e.g. coordinates of other dynamic objects) or stop on (e.g. railway tracks).

\begin{table}%
\begin{tabular}{l|rcl}
Vehicle speed & $v_0$ &$\leqslant$& $45\,\unit{m/s}$ ($\approx 160\,\unit{km/h}, 100\,\unit{mph}$)\\\hline
Braking decelerations & $a$ &$\in$& $[-9\,\unit{m/s^2}, -1\,\unit{m/s^2}]$\\\hline
Planning horizon & $t\subs{hzn}$ &$=$& $10\,\unit{s}$\\\hline
Replanning interval & $\Delta t\subs{plan}$ &$=$& $0.25\,\unit{s}$\\\hline
Valve speed & $\kappa$ &$=$& $100\,\unit{m\,s^{-3}}$\\\hline
Valve transition time  & $\Delta t\subs{valve}$ &$\leqslant$& $0.08\,\unit{s} = (a_{\max} - a_{\min})/\kappa$
\end{tabular}
\caption{Numerical reference values of relevant parameters introduced throughout the paper, used only where explicitly indicated and exclusively to provide realistic orders of magnitude, without loss of generality.}
\vspace{-5pt}
\label{tab:referencevalues}
\end{table}

\subsection{A Single Stopping Trajectory}

The basic element of the emergency stopping problem description is a timing $\sigma(t,v_0, t\subs{fail}, a)$ which drives at constant speed $v_0$ until $t\subs{fail}$, and then decelerates with some negative $a$ until it comes to a halt.\footnote{Note that we strictly use $t\subs{fail}$ as the instant when the emergency deceleration engages. Any deterministic delay between actual failure and deceleration onset, such as by hydraulics, brake pad motion or signal times that can be determined \emph{a-priori}, is considered included.} 
This timing is given by 
\begin{equation}
\sigma(t,v_0, t\subs{fail},a) = \begin{cases}
v_0 \, t & \text{for $t \leqslant t\subs{fail}$}\\
s_\blacksquare & \text{for $t \geqslant t_\blacksquare$}\\
v_0 \, t\subs{fail} + \frac{a \, (t-t\subs{fail})^2}{2} & \text{else,}\\
\end{cases}
\end{equation}
where the stopping time and distance are given by
\begin{equation}
t_\blacksquare = \phantom{v_0\,}t\subs{fail} - v_0 / a\quad\text{and}\quad
s_\blacksquare = v_0\, t\subs{fail} - v_0^2 / 2 a.
\end{equation}

With the trajectory shape completely specified by parameters $v_0, t\subs{fail},a$, the Euler--Lagrange form \eqref{eq:euler-lagrange} of the penalty cost functional can be expressed as a penalty \emph{function}:
\begin{equation}
P(v_0, t\subs{fail},a)
:= \mathcal P[\sigma(\,\cdot\,,\, v_0, t\subs{fail},a)] = \!\!\int_{t\subs{now}}^{t\subs{hzn}} \!\!\!\!\!\! \dee t\, W(t, \sigma(t, ...)).\label{eq:penaltyfunction}
\end{equation}
Thereby, the optimization simplifies to $a^* \in \argmin_{a} P(a)$.

\subsection{Adaptation to the Actual Problem}

The actual planning problem, however, is more complex than optimizing $P$ for $a$. At each planning instant of the vehicle (i.e. strictly \emph{before} the emergency), $v_0$ is known, but $t\subs{fail}$ is not: The system may fail at any time within the planning interval $T\subs{plan} = [t\subs{now}, t\subs{plan})$.\footnote{Do note that it may (and typically should) not fail within $T\subs{plan}$ at all; however, since any planned action is only executed \emph{iff} the system fails, the failure within $T\subs{plan}$ acts as a stochastic precondition in the planning. Thereby, all modeling is independent of $p(\text{fail})$, which would typically be difficult to determine.} With a relatively short $\Delta t\subs{plan}$ (cf. Tab.~\ref{tab:referencevalues}), it is considered unlikely that there is considerable prior knowledge about \emph{when} $t\subs{fail}$ would occur within $T\subs{plan}$, so we assume a uniform $t\subs{fail} \sim \mathcal U(T\subs{plan})$.

Along with the unknown $t\subs{fail}$, even $a$ is unknown: Since $a$ is a mechanical parameter (namely the state of valve \textbf{C} in Fig.~\ref{fig:hydraulic-electronic}), it cannot be switched instantaneously. Instead, if the optimal valve state from the previous planning cycle was $a\subs{prev}$, and our (yet undefined) optimization process obtains $a\subs{next}$ as the next optimal solution, the valve will take some non-negligible interval $[t\subs{now}, t\subs{valve}]$ to transition, modeled linearly as $t\subs{valve} = t\subs{now} + \kappa\,(a\subs{next}-a\subs{prev})$ using a signed ``valve speed'' $\kappa$ with $\operatorname{sign}\kappa = \operatorname{sign}(a\subs{next}-a\subs{prev})$. For example, for values as in Tab.~\ref{tab:referencevalues}, the probability of failure during valve transition can be up to $32\,\%$. We note that $a \not\sim \mathcal U(T\subs{plan})$: If valve motion is approximately linear with $a$, but $t\subs{valve} < \Delta t\subs{plan}$, $a$ is uniformly distributed during \smash{$[t\subs{now}, t\subs{valve}]$}, but constant thereafter, with $t\subs{valve}$ depending on $a\subs{prev} - a\subs{next}$. We therefore denote the unknown value as $\alpha(a\subs{prev}, a\subs{next}, t\subs{fail})$ and find the actual curve family to be
\begin{equation}
\sigma(t,v_0, t\subs{fail},\alpha(a\subs{prev}, a\subs{next}, t\subs{fail})).
\end{equation}

The complete planning problem thus presents itself as minimizing the expected penalty value for a choice of $a\subs{next}$
\begin{equation}
	a^* \in \argmin_{a\subs{next}}\; \tfrac{1}{\Delta t\subs{plan}} \langle P(..., a\subs{next}, ...)\rangle,\label{eq:planningproblem}
\end{equation}
using $\langle\,\cdot\,\rangle$ as the non-normalized expected value over $t\subs{fail}$,
\begin{equation}
	 \langle P(..., a\subs{next}, ...) \rangle := \!\! \int_{t\subs{now}}^{t\subs{plan}}\!\!\!\!\!\!\!\! \dee t\subs{fail}\; P(v_0, t\subs{fail},\alpha(a\subs{prev}, a\subs{next}, t\subs{fail})).
	\label{eq:expected-value}
\end{equation}

For its solution, we note that $P$ depends on $W(t,s)$ (the risk predictions), which is typically not analytic but an array, so \emph{analytic} solving is not feasible. \emph{Iterative} solvers have difficulty guaranteeing time or quality constraints (cf.~\cite{RZ.2015b}), and for realtime safety applications, we require both. This points to discretization and subsequent global optimization of the discretized set, yet a direct, exhaustive solution of \eqref{eq:planningproblem} over $a\subs{next}$ is prohibitive for lightweight realtime applications: To test any $a$, we must evaluate all $t\subs{fail}$, and each $t\subs{fail}$ yields a timing $\sigma(t)$ to be integrated over (as in \eqref{eq:penaltyfunction}) to evaluate its cumulative penalty costs $P$. Hence, we are looking for ways to simplify the computation by reusing computations, not between planning cycles (we consider each planning step a new problem), but within a single cycle.

\section{PROBLEM SIMPLIFICATION}\label{sec:simplification}

\begin{figure}%
\vspace{5pt}
\hspace{14pt}
\begin{overpic}[
]{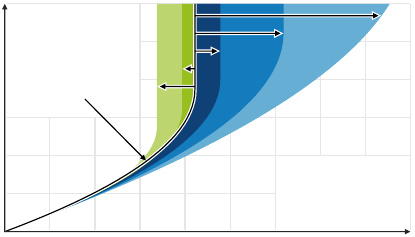}%

\put(93, 52){\footnotesize $-2\,\unit{\frac{m}{s^2}}$}
\put(70, 48){\footnotesize $-3\,\unit{\frac{m}{s^2}}$}
\put(54, 44){\footnotesize \color{white} $-4\,\unit{\frac{m}{s^2}}$}
\put(34, 40){\footnotesize  $-5\,\unit{\frac{m}{s^2}}$}
\put(28, 35){\footnotesize  $-6\,\unit{\frac{m}{s^2}}$}

\put(-1,-3){\footnotesize $0\,\unit{m}$}
\put(20,-3){\footnotesize $20\,\unit{m}$}
\put(42,-3){\footnotesize $40\,\unit{m}$}
\put(64,-3){\footnotesize $60\,\unit{m}$}
\put(85,-3){\footnotesize $80\,\unit{m}$}
\put(100,-3){\small $s$}
\put(-15,1){\parbox{1cm}{\raggedleft\footnotesize $0\,\unit{s}$}}
\put(-15,9){\parbox{1cm}{\raggedleft\footnotesize $1\,\unit{s}$}}
\put(-15,19){\parbox{1cm}{\raggedleft\footnotesize $2\,\unit{s}$}}
\put(-15,28){\parbox{1cm}{\raggedleft\footnotesize $3\,\unit{s}$}}
\put(-15,37){\parbox{1cm}{\raggedleft\footnotesize $4\,\unit{s}$}}
\put(-15,46){\parbox{1cm}{\raggedleft\footnotesize $5\,\unit{s}$}}
\put(-15,55){\parbox{1cm}{\raggedleft\footnotesize $6\,\unit{s}$}}

\put(72, 9){\parbox{1.4cm}{
\raggedright\footnotesize
\begin{align*}v_0 &= 15\,\unit{\tfrac{m}{s}}\\
a\subs{prev} &= -4\,\unit{\tfrac{m}{s^2}}
\end{align*}
}}

\put(4, 42){\parbox{1.6cm}{
\raggedright\footnotesize
the timing for $t\subs{fail}{=}t\subs{now}$, $\sigma(t, t\subs{now}, ...)$, always brakes with $a\subs{prev}$
}}
\end{overpic}
\vspace{5pt}
\caption{Areas covered for different choices of $a\subs{next}$ overlap, starting with the timing $\sigma(t, t\subs{now}, ...)$. This motivates the attempt to use an antiderivative of $W(t,s)$ w.r.t $s$ to determine the expected value of penalties accumulated within the area for a particular choice of $a\subs{next}$.}%
\label{fig:flaechenintegrale}%
\end{figure}
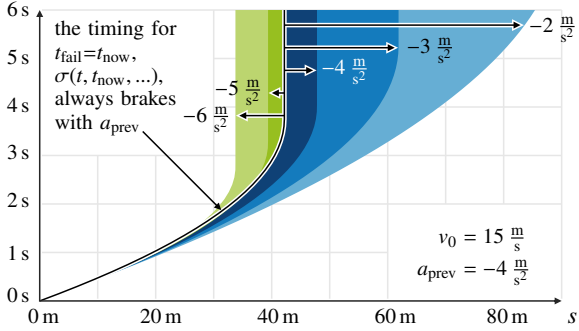

First, we note that at the beginning of each planning cycle, the following parameters are known: The current vehicle speed $v_0$, the current valve state $a\subs{prev}$, as well as the parametric constants $t\subs{now}$, $t\subs{plan}$ and $t\subs{hzn}$. Optimization result is $a\subs{next}$, whereas $t\subs{fail} \in [t\subs{now}, t\subs{now}+\Delta t\subs{plan}]$ always remains unknown. We introduce the three-parametric shorthand
\begin{align}
\sigma(t, t\subs{fail}, a\subs{next}) &= \sigma(t, v_0, t\subs{fail}, \alpha(\alpha\subs{prev}, \alpha\subs{next}, t\subs{fail})),
\end{align}
and state the goal to establish some \emph{easily precomputed} $\mathcal I$ (an antiderivative w.r.t. $s$), to achieve a form
\begin{equation}
\begin{aligned}
 \langle P(..., a\subs{next}, ...) \rangle = \!\!\int_{t\subs{now}}^{t\subs{hzn}} \!\dee t \; \biggl(\;
&\mathcal I(t, \sigma(t, \underbrace{t\subs{plan}}_{\text{\clap{range over possible $t\subs{fail}$}}}, a\subs{next}))\\[-2pt]
-
\;&\mathcal I(t, \sigma(t, \overbrace{t\subs{now}}, a\subs{next}))\;\biggr)
\end{aligned}
\label{eq:integralbild}
\end{equation}
such that \textbf{(a)} time steps up to $t\subs{hzn}$ can be treated independently, and \textbf{(b)} to evaluate one candidate $a\subs{next}$, we must no longer integrate over all timings (= trajectories) for all possible $t\subs{fail}$, but instead look up in the precomputed $\mathcal I$. 
The first condition \textbf{(a)} can be readily rearranged by
\begin{align}
 \langle P(..., a\subs{next}, ...) \rangle = & \int_{t\subs{now}}^{t\subs{plan}}\!\!\!\!\! \dee t\subs{fail}  \int_{t\subs{now}}^{t\subs{hzn}}\!\!\!\!\! \dee t\; W(t, \sigma(t, t\subs{fail}, a\subs{next}))\\
\intertext{where we may apply Fubini's theorem to obtain}
= & \int_{t\subs{now}}^{t\subs{hzn}}\!\!\!\!\! \dee t \int_{t\subs{now}}^{t\subs{plan}}\!\!\!\!\! \dee t\subs{fail} \; W(t, \sigma(t, t\subs{fail}, a\subs{next})),\label{eq:fubini}
\end{align}
since all physically possible timings are necessarily continuous and all intervals are closed. 
%
For \textbf{(b)}, we seek some 
$
\mathcal I(t, \sigma(t, t\subs{fail}, a\subs{next}))
$
that is easily precomputed (``pre'' in the sense of \emph{before} picking any candidate $a\subs{next}$), so we must integrate over $s$ instead of $t\subs{fail}$, since evaluating $\sigma(t, t\subs{fail}, a\subs{next})$ requires $a\subs{next}$. Hence we mean to establish a substitution function $\tau\subs{fail}$ s.t. for any valid arc length $s$
\begin{equation}
\sigma(t, \tau\subs{fail}(t, s), a\subs{next}) = s\quad\text{and thus}
\end{equation}
\begin{equation}
\!\;\;\int_{t\subs{now}}^{t\subs{plan}}\!\!\!\!\! \dee t\subs{fail} \; W(t, \sigma(t, t\subs{fail}, a\subs{next}))
%
 =\!\!  \int_{\sigma(t, t\subs{now}, a\subs{next})}^{\sigma(t, t\subs{plan}, a\subs{next})}\!\!\!\!\!\!\!\!\!\!\!\!\!\!\!\!\!\!\!\! \dee s \; W(t, s) \left.\frac{\partial \tau\subs{fail}}{\partial s}\right|_{t,s}
\end{equation}
\begin{equation}
 =:  \;\; \mathcal I(t, \sigma(t, t\subs{plan}, a\subs{next}))
- \mathcal I(t, \sigma(t, t\subs{now}, a\subs{next}))
\end{equation}
which allows to specify $\mathcal I$ as required in \eqref{eq:integralbild} as
\begin{equation}
\mathcal I(t,s) = \int_0^s\!\dee s\; W(t,s)  \left.\frac{\partial \tau\subs{fail}}{\partial s}\right|_{t,s}
\label{eq:integrabild-multiplikation}
\end{equation}

\begin{figure}%
\vspace{4pt}
\vspace{5pt}
\hspace{14pt}
\begin{overpic}[
]{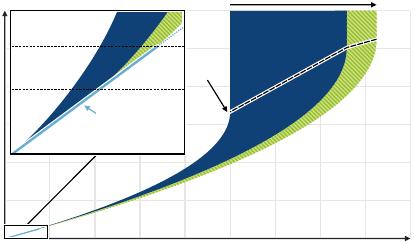}%

\put(13,23.5){\footnotesize exaggerated detail}

\put(68,59){\footnotesize$t\subs{fail}$}

\put(67,33){\footnotesize\color{white} ${\mathbf{B}}_\blacktriangleright$}
\put(65,47){\footnotesize\color{white} ${\mathbf{B}}_\blacksquare$}

\put(84,43){\footnotesize ${\mathbf{C}}_\blacktriangleright$}
\put(85,51){\footnotesize ${\mathbf{C}}_\blacksquare$}

\put(40,52){\footnotesize ${\mathbf{C}}$}
\put(32,52){\footnotesize\small\color{white} ${\mathbf{B}}$}
\put(24,30){\footnotesize $\text{\textbf{A}}$}

\put(47,42){\small $\left[\!\begin{smallmatrix}s_\blacksquare\\t_\blacksquare\end{smallmatrix}\!\right]$}

\put(5,39){\small $t\subs{valve}$}
\put(5,50){\small $t\subs{plan}$}
\put(-1,-3){\footnotesize $0\,\unit{m}$}
\put(20,-3){\footnotesize $20\,\unit{m}$}
\put(42,-3){\footnotesize $40\,\unit{m}$}
\put(64,-3){\footnotesize $60\,\unit{m}$}
\put(85,-3){\footnotesize $80\,\unit{m}$}
\put(100,-3){\small $s$}
\put(-15,1){\parbox{1cm}{\raggedleft\footnotesize $0\,\unit{s}$}}
\put(-15,9){\parbox{1cm}{\raggedleft\footnotesize $1\,\unit{s}$}}
\put(-15,19){\parbox{1cm}{\raggedleft\footnotesize $2\,\unit{s}$}}
\put(-15,28){\parbox{1cm}{\raggedleft\footnotesize $3\,\unit{s}$}}
\put(-15,37){\parbox{1cm}{\raggedleft\footnotesize $4\,\unit{s}$}}
\put(-15,46){\parbox{1cm}{\raggedleft\footnotesize $5\,\unit{s}$}}
\put(-15,55){\parbox{1cm}{\raggedleft\footnotesize $6\,\unit{s}$}}
\end{overpic}
\vspace{10pt}
\caption{The set of all possible sub-trajectories for a given transition is composed as follows: The set \textbf{A} of sub-trajectories that have not failed yet (thin light blue line); the set \textbf{B} of sub-trajectories that ensue for failures at some $t\subs{fail} \in [t\subs{now}, t\subs{valve})$ which brake with intermediate accelerations $[a\subs{prev}, a\subs{now})$; and the set \textbf{C} of sub-trajectories for failures $t\subs{fail} \in [t\subs{valve}, t\subs{plan}]$ that brake with the target deceleration of $a\subs{next}$ (hatched green). The latter two we further distinguish into ${\mathbf{B}}_\blacktriangleright$, $\mathbf{B}_\blacksquare$ and  $\mathbf{C}_\blacktriangleright$, ${\mathbf{C}}_\blacksquare$ by whether the vehicle already has stopped (indicated by the dashed $[s_\blacksquare, t_\blacksquare]\transp$ line).
}%
\label{fig:trajseparation}%
\end{figure}

To compute \eqref{eq:integrabild-multiplikation}, we require $\partial \tau\subs{fail}/\partial s$, intuitively the density of trajectories passing through some space segment by change in failure times. We distinguish between the following sets of sub-trajectories, as shown in Fig.~\ref{fig:trajseparation}:
\begin{itemize}
	\item Sub-trajectories \textbf{A} that have not failed yet and hence lie on the regularly planned trajectory. All trajectories start in this set at the instant $t=t\subs{now}$ with the common point $\sigma(t\subs{now}) = 0$, but branch off to a different set ($\mathbf{B}$ or $\mathbf{C}$) once they fail.	Since the entire planning process is conditioned upon the assumption that failure is \emph{certain} within $[t\subs{now}, t\subs{plan}]$, the longest sub-trajectory in this set lasts until $t\subs{fail}$, when it is the last to fail and branch off. The set \textbf{A} is special in that all sub-trajectories therein overlap perfectly, but their density decreases with $t$.
	\item Sub-trajectories $\mathbf{B}$ that have failed, but the valve had not yet reached $a\subs{next}$. These sub-trajectories decelerate with varying decelerations $[a\subs{prev}, a\subs{next}]$. As seen in Fig.~\ref{fig:trajseparation}, these sub-trajectories can cover a wide interval over $s$ for large $|a\subs{next}-a\subs{prev}|$. If $a\subs{next} = a\subs{prev}$, \textbf{B} is empty.
	\item Sub-trajectories $\mathbf{C}$ that have failed, and the valve did reach $a\subs{next}$ before that. These sub-trajectories all decelerate with $a\subs{next}$, and are only spaced by the vehicle driving along its original path for a longer time.
\end{itemize}

\begin{figure}%
\vspace{4pt}
\vspace{7pt}
\hspace{14pt}
\begin{overpic}[
]{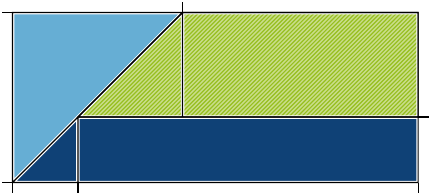}%

\put(12.5,6){\footnotesize\color{white} ${\mathbf{B}}_1$}
\put(56,9){\footnotesize\color{white}    ${\mathbf{B}}_2$}
\put(34,23.5){\footnotesize  ${\mathbf{C}}_1$}
\put(68.25,28.5){\footnotesize  ${\mathbf{C}}_2$}

\put(10,30){\parbox{2cm}{\footnotesize$t \leqslant t\subs{fail}$\vspace{5pt}\newline \textbf{A}}}

\put(-1,-3){\small $t\subs{now}$}
\put(13,-3){\small $t\subs{valve}$}
\put(95,-3){\small $t\subs{hzn}$}
\put(45,-3){\small $t$}
\put(-7,2){\small $t\subs{now}$}
\put(-7,42){\small $t\subs{plan}$}
\put(-7,23){\small $t\subs{fail}$}
\put(39,47){\small $t\subs{plan}$}
\put(100,17){\small $t\subs{valve}$}
\end{overpic}
\vspace{10pt}
\caption{We distinguish between trajectories which have not failed/decelerated yet (\textbf{A}, solid light blue), trajectories where the valve was activated in an intermediate state ($\mathbf{B}$, solid dark blue), and trajectories where the valve was activated in its constant state $a\subs{next}$ ($\mathbf{C}$, hatched green). The subsets $\mathbf{B}_1$, $\mathbf{B}_2$ and $\mathbf{C}_1$, $\mathbf{C}_2$ defined to obtain simple boundaries for integration in \eqref{eq:area-p-a}--\eqref{eq:area-p-c}.}%
\label{fig:trajseparation-integration}%
\end{figure}
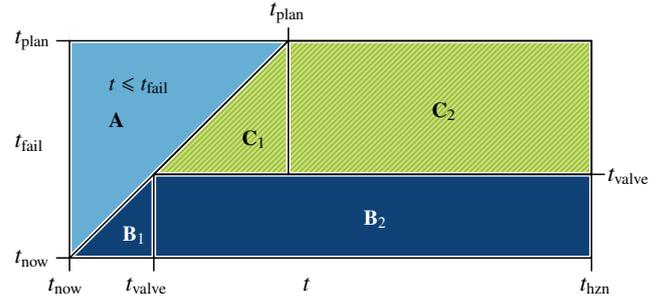

The sub-trajectories within these sets are given by
\begin{align}
\sigma^{\mathbf{A}}(t, t\subs{fail}) &= v_0\, t\;,\label{eq:sigma-A}\\[5pt]
\sigma^{\mathbf{B}}(t, t\subs{fail}) &=
\begin{cases}
	v_0 t + \frac{1}{2} (a\subs{prev} + \kappa t\subs{fail}) (t-t\subs{fail})^2 & t < t_\blacksquare\\
	v_0   t\subs{fail} - v_0^2 / (2 (a\subs{prev} + \kappa t\subs{fail})) & \text{else, and}\!\!\!
\end{cases}\label{eq:sigma-B}\\[5pt]
\sigma^{\mathbf{C}}(t, t\subs{fail}) &=
\begin{cases}
	v_0 t + \frac{1}{2} a\subs{next} (t-t\subs{fail})^2 \hspace{35pt}& t < t_\blacksquare\\
	v_0 t\subs{fail} - v_0^2 / (2 a\subs{next}) & \text{else.}
\end{cases}\label{eq:sigma-C}
\end{align}
\newcommand{\hneg}{\!\!\!\!\!\!\!\!}
Using this distinction, we state for the expected penalty costs
\begin{equation}
 \langle P(..., a\subs{next}, ...) \rangle = 
\langle P\rangle^{\mathbf{A}} + \langle P \rangle^{\mathbf{B}} + \langle P \rangle^{\mathbf{C}}
\end{equation}
where $\langle P \rangle^{\mathbf{A}}$, ..., $\langle P \rangle^{\mathbf{C}}$ are the expected penalty costs for sub-trajectories within $\mathbf A$ through $\mathbf C$, namely, by the boundaries of integration shown in Fig.~\ref{fig:trajseparation-integration},
\begin{align}
\langle P\rangle^{\mathbf{A}} &\!\!\,=\!\!\!
\int_{t\subs{now}}^{t\subs{plan}}\hneg \dee t \int_{t}^{t\subs{plan}} \hneg\dee t\subs{fail} W(t, ...) \;=\! \int_{t\subs{now}}^{t\subs{plan}}\hneg\dee t\; (t\subs{plan}{-}t) \, W(t, v_0t)\label{eq:area-p-a}\\[5pt]
\langle P\rangle^{\mathbf{B}} &\!\!\,=\!\!\!
\underbrace{\int_{t\subs{now}}^{t\subs{valve}}\hneg \dee t \int_{t\subs{now}}^{t}\hneg\;\; \dee t\subs{fail} \; W(t, ...)}_{\mathbf B_1}
+
\underbrace{\int_{t\subs{valve}}^{t\subs{hzn}}\hneg\;\; \dee t \int_{t\subs{now}}^{t\subs{valve}} \hneg \dee t\subs{fail} W(t, ...)}_{\mathbf B_2}\label{eq:area-p-b}\\
\langle P\rangle^{\mathbf{C}} &\!\!\,=\!\!\!
\underbrace{\int_{t\subs{valve}}^{t\subs{plan}}\hneg \dee t \int_{t\subs{valve}}^{t}\hneg\;\; \dee t\subs{fail} \; W(t, ...)}_{\mathbf C_1}
+
\underbrace{\int_{t\subs{plan}}^{t\subs{hzn}}\hneg\;\; \dee t \int_{t\subs{valve}}^{t\subs{plan}} \hneg \dee t\subs{fail} W(t, ...)}_{\mathbf C_2}\label{eq:area-p-c}
\end{align}
where the ellipses (``...'') denote $\sigma^{\mathbf{A}}(t, t\subs{fail})$ through $\sigma^{\mathbf{C}}(t, t\subs{fail})$ from the previous \eqref{eq:sigma-A}--\eqref{eq:sigma-C}  respectively, such that
\begin{equation}
 \langle P(..., a\subs{next}, ...) \rangle =  \langle P\rangle^{\mathbf{A}}  +  \langle P \rangle^{\mathbf{B}} +  \langle P\rangle^{\mathbf{C}}.
\end{equation}
We note that sub-trajectories contained in $\mathbf{A}$ lie on the regularly planned trajectory and range up to $t\subs{fail}$---thus, neither their shape nor their density is affected by the choice of $a\subs{next}$. Therefore, $\langle P\rangle^{\mathbf{A}}$ is a constant term in the optimization that does not affect the solution $a^*$. We thus need not specify a substitution function for $\langle P\rangle^{\mathbf{A}}$ to obtain $a^*$ in \eqref{eq:planningproblem}.\footnote{Also note that, since we aimed to pose the emergency planning problem as consistent with the regular maneuver planning, which minimizes the penalty costs of the regular trajectory, $\langle P\rangle^{\mathbf{A}}$ should typically be very low.} For $\langle P\rangle^{\mathbf{B}}$ and $\langle P\rangle^{\mathbf{C}}$, we define the substituted integrals via limits, to avoid integrating over discontinuous boundaries. We define 
\begin{align}
\langle P\rangle^{\mathbf{B}}_{\boldsymbol\varepsilon} &=
\underbrace
{
\int_{\varepsilon_1}^{t\subs{valve}}\hneg\;\;\,\dee t\quad
\int_{\sigma^{\mathbf{B}}(t, t\subs{now})}^{\sigma^{\mathbf{B}}(t, t-\varepsilon_2)}\hneg\dee s\;
W(t,s)\left.\frac{\partial \tau^{\mathbf{B}}}{\partial s}\right|_{t,s}
}_{\mathbf B_1}\notag\\
&+
\underbrace
{
\int_{t\subs{valve}+\varepsilon_5}^{t\subs{hzn}}\hneg\,\dee t\quad
\int_{\sigma^{\mathbf{B}}(t, t\subs{now})}^{\sigma^{\mathbf{B}}(t, t\subs{valve})}\hneg\dee s\;
W(t,s)\left.\frac{\partial \tau^{\mathbf{B}}}{\partial s}\right|_{t,s}
}_{\mathbf B_2}\label{eq:area-p-b-tau}\\
\mathllap{\text{and}}\quad\langle P\rangle^{\mathbf{C}}_{\boldsymbol\varepsilon} &=
\underbrace
{
\int_{t\subs{valve}+\varepsilon_3}^{t\subs{plan}}\hneg\dee t\quad
\int_{\sigma^{\mathbf{C}}(t, t\subs{valve})}^{\sigma^{\mathbf{B}}(t, t-\varepsilon_4)}\hneg\dee s\;
W(t,s)\left.\frac{\partial \tau^{\mathbf{C}}}{\partial s}\right|_{t,s}
}_{\mathbf C_1}\notag\\
&+
\underbrace
{
\int_{t\subs{plan}+\varepsilon_6}^{t\subs{hzn}}\hneg\;\,\dee t\quad
\int_{\sigma^{\mathbf{C}}(t, t\subs{valve})}^{\sigma^{\mathbf{B}}(t, t\subs{plan})}\hneg\dee s\;
W(t,s)\left.\frac{\partial \tau^{\mathbf{C}}}{\partial s}\right|_{t,s}
}_{\mathbf C_2}\label{eq:area-p-c-tau}\\
\mathllap{\text{such that}}\quad&\quad
 \langle P \rangle -  \langle P\rangle^{\mathbf{A}}  = 
\lim_{\boldsymbol\varepsilon \rightarrow \boldsymbol 0}\;\;
\langle P \rangle^{\mathbf{B}}_{\boldsymbol\varepsilon} \;+\;  \langle P\rangle^{\mathbf{C}}_{\boldsymbol\varepsilon}
\end{align}
under $\varepsilon_1 \geqslant \varepsilon_2$ and $\varepsilon_3 \geqslant \varepsilon_4$. The existence of this limit is shown in \cite{fabianma}. To derive the substitution functions $\tau^{\mathbf{B}}$ and $\tau^{\mathbf{C}}$, we distinguish the sets $\mathbf{B}$ and $\mathbf{C}$ further into regions while the vehicle still moves, and regions where it already stopped, namely ${\mathbf{B}} = {\mathbf{B}}_\blacktriangleright \;\dot\cup\; {\mathbf{B}}_\blacksquare$ and ${\mathbf{C}} = {\mathbf{C}}_\blacktriangleright \;\dot\cup\; {\mathbf{C}}_\blacksquare$, cf. Fig.~\ref{fig:trajseparation}.

\subsubsection{$\mathbf{B}_\blacktriangleright$---Vehicle decelerating, valve stopped in transition}

In this case, we find for the trajectories
\begin{equation}
\sigma^{\mathbf{B}}(t,v_0, t\subs{fail},a\subs{prev}) = v_0\, t + \tfrac{1}{2} (a\subs{prev} + \kappa\,t\subs{fail}) (t-t\subs{fail})^2
\end{equation}
\begin{equation}
= \frac{\kappa t\subs{fail}^3}{2}
+
\left(\frac{a\subs{prev}}{2}-\kappa t\right) t\subs{fail}^2
+
\left(\frac{\kappa t\mathrlap{{}^2}}{2}-a\subs{prev}t\right) t\subs{fail}
+
v_0t + \frac{a\subs{prev} t^2}{2},
\end{equation}
where the absolute value of the third-order term is, in the exemplary quantities of Tab.~\ref{tab:referencevalues}, bounded by \smash{$\left|(\kappa/2) \,t\subs{valve}^3\right| < 2.6\,\unit{cm}$}.\footnote{Since in the case of the valve stopping in transition we have $t\subs{fail} \leqslant t\subs{valve}$.} As this will typically be considerably lower than positional uncertainty of the predictions, we approximate $s$ by omitting this term to proceed with
\begin{equation}
%
\tilde \sigma^{\mathbf{B}}(...) = 
\left(\frac{a\subs{prev}}{2}-\kappa t\right) t\subs{fail}^2
+
\left(\frac{\kappa t\mathrlap{{}^2}}{2}-a\subs{prev}t\right) t\subs{fail}
+
v_0t + \frac{a\subs{prev} t^2}{2}
\end{equation}
\begin{equation}
\tau\subs{fail}^{\mathbf{B}}(t,s) = -\frac{\beta_{\blacktriangleright} \pm \operatorname{sign}(\kappa)}{2\,\alpha_{\blacktriangleright}} \, \sqrt{\beta_{\blacktriangleright}^2 - 4\,\alpha_{\blacktriangleright}\,\gamma_{\blacktriangleright} + 4\,\alpha_{\blacktriangleright}\,s}
\end{equation}
with $\alpha_{\blacktriangleright} = \tfrac{1}{2} a\subs{prev} - \kappa t$, $\beta_{\blacktriangleright} = \tfrac{1}{2} \kappa t^2 - a\subs{prev}t$ and $\gamma_{\blacktriangleright} = v_0t + \tfrac{1}{2}a\subs{prev}t^2$. It can be seen that $\tau\subs{fail}^{\mathbf{B}}(t,s)$ is not uniquely determined at $s$. Particularly we find that
\begin{equation}
\biggl. \frac{\partial \tilde{\sigma}^\mathbf{B}}{\partial t\subs{fail}} \biggr|_{t\subs{fail}, t} = 0 \Leftrightarrow t\subs{fail} = t + \frac{3 \, \kappa\, t^2}{2\,a\subs{prev}-4 \, \kappa\, t}
\end{equation}
where arc length $s$ reaches an extremum over $t\subs{fail}$ and several arc lengths may occur twice along $t\subs{fail}$ (details in \cite{fabianma}).

%
%
%

\subsubsection{$\mathbf{B}_\blacksquare$---Vehicle at rest, valve stopped in transition}

In this case we have the trajectories
\begin{equation}
\sigma^{\mathbf{B}}(t,t\subs{fail}) = v_0 \, t\subs{fail} - v_0^2/(2\, a\subs{prev} + 2\,\kappa\,t\subs{fail})
\end{equation}
which gives
\begin{equation}
\tau\subs{fail}^{\mathbf{B}, \pm}(t,s) = \frac{-\beta_{\blacksquare} \pm \operatorname{sign}(\kappa)}{2\,\alpha_{\blacksquare}} \, \sqrt{\beta_{\blacksquare}^2 - 4\,\alpha_{\blacksquare} \, \gamma_{\blacksquare}}
\end{equation}
with $\alpha_{\blacksquare} = 2v_0\kappa$, $\beta_{\blacksquare} = 2v_0a\subs{prev}-2\kappa s$ and $\gamma_{\blacksquare} = -v_0^2-2\alpha_0s$. Again looking for singular points gives the solutions
\begin{equation}
\biggl. \frac{\partial \sigma^{\mathbf{B}}}{\partial t\subs{fail}} \biggr|_{t\subs{fail}, t} = 0 \Leftrightarrow t\subs{fail}^\pm = \frac{-a\subs{prev} \pm \sqrt{-\tfrac{1}{2}\,\kappa\,v_0}}{\kappa},
\end{equation}
for $\kappa < 0$, of which only \smash{$t\subs{fail}^+ = (-a\subs{prev} + \sqrt{...})/\kappa$} can lie within $[t\subs{now}, t\subs{valve}]$, and only for specific parameters:
\begin{equation}
t\subs{fail}^+ \in [t\subs{now}, t\subs{valve}] \Leftrightarrow v_0 < -2\,a^2\subs{next} / \kappa
\end{equation}
By Tab.~\ref{tab:referencevalues}, this condition is satisfied only for $v_0 \leqslant 1.62\,\unit{m/s}$; under relevant conditions, $\sigma^{\mathbf{B}}$ is non-singular w.r.t. $t\subs{fail}$.

%

\subsubsection{$\mathbf{C}_\blacktriangleright$---Vehicle decelerating, valve reached $a\subs{next}$}

In this case we have the trajectories
\begin{equation}
\sigma^{\mathbf{C}}(t, t\subs{fail}) = v_0 \, t + \tfrac{1}{2} \, a\subs{next} \, (t-t\subs{fail})^2,
\end{equation}
\begin{equation}
\text{and}\quad \tau\subs{fail}^{\mathbf{C},\pm} (t,s) = \frac{a\subs{next}\,t \pm \sqrt{-2\,a\subs{next}\,v_0\,t+2\,a\subs{next}\,s}}{a\subs{next}}
\end{equation}
which is strictly $\tau\subs{fail}^{\mathbf{C}} = \tau\subs{fail}^{\mathbf{C},+}$ for $t > t\subs{fail}$.

\subsubsection{$\mathbf{C}_\blacksquare$---Vehicle at rest, valve reached $a\subs{next}$}

This final case contains the trajectories
\begin{equation}
\sigma^{\mathbf{C}}(t,t\subs{fail}) = v_0 \, t\subs{fail} - v_0^2/(2\,a\subs{next})
\end{equation}
\begin{equation}
\text{with}\quad\tau\subs{fail}^{\mathbf{C}}(t,s) = s/v_0 + v_0/(2\,a\subs{next}).
\end{equation}

\subsubsection{Partial Derivatives of $\tau\subs{fail}(t,s)$ over $\mathbf{B}\, \dot\cup\,\mathbf{C}$}

The results of the previous sections give the partial derivatives of $\tau\subs{fail}$ (as required in \eqref{eq:area-p-b-tau} and \eqref{eq:area-p-c-tau}) as
\begin{align}
\left.\frac{\partial \tau\subs{fail}^{\mathbf{B}}}{\partial s}\right|_{t,s} &= \begin{cases}
\displaystyle
\frac{\operatorname{sign}(\kappa)}{\sqrt{\beta_{\blacktriangleright} - 4\alpha_{\blacktriangleright}\gamma_{\blacktriangleright}+4\alpha_{\blacktriangleright} s}}
& s > s_{\blacksquare}\\[20pt]
\displaystyle
\frac{1}{2\,v_0} - \frac{\operatorname{sign}(\kappa) (\beta_{\blacksquare}+4\,\kappa\,s)}{2v_0\sqrt{\beta_{\blacksquare}^2-4\alpha_{\blacksquare}\gamma_{\blacksquare}}} & \text{else, and}
\end{cases}\label{eq:dtau-b-ds}\\[3pt]
%
%
\left.\frac{\partial \tau\subs{fail}^{\mathbf{C}}}{\partial s}\right|_{t,s} &= \begin{cases}
\displaystyle
\frac{1}{\sqrt{-a\subs{next}} \sqrt{2 v_0 t + 2 s}}
\hspace{14pt}
& s > s_{\blacksquare}\\[10pt]
\displaystyle
v_0^{-1}
& \text{else.}
\end{cases}
\label{eq:dtau-c-ds}
\end{align}

\section{PROBLEM SOLUTION}\label{sec:problem-solution}

Having established $\partial \tau/\partial s$ now allows us to compute $\mathcal I(t,s)$ by \eqref{eq:integrabild-multiplikation}, to approximately optimize $ \langle P(..., a\subs{next}, ...) \rangle$ in \eqref{eq:planningproblem}.

\subsection{Discretization}

We use the discretizations $\hat{\Delta}t$ for prediction time and $\hat{\Delta}s$ for stopping distance to define the following sets:
\begin{align}
\hat T &= \Set{ \hat{t} | \hat{t} = t\subs{now} + m \; \hat{\Delta}t, \;  \hat{t} \leqslant t\subs{hzn},\; m \in \{0, 1, 2, ...\}  }\\
\hat S &= \Set{ \hat{s} | \hat{s} = 0\,\unit{m} + n \; \hat{\Delta}s, \; n \in \{0, 1, 2, ...\}  }\\
\hat S(t) &= \Set{ s | s \in \hat S, \; \sigma\subs{min}(t) \leqslant s \leqslant \sigma\subs{max}(t)}\\
\hat A &= \Set{ a\subs{min},\; a\subs{min}+\Delta a,\; ...,\; a\subs{max}-\Delta a,\; a\subs{max} }
\end{align}
where $\sigma\subs{min}(t)$ and $\sigma\subs{max}(t)$ are the shortest and longest possible stopping trajectories respectively,
\begin{align}
\sigma\subs{min}(t) &= \sigma(t,\; t\subs{fail} = (a\subs{max}-a\subs{prev})/\kappa,\; a\subs{max}) \\
\text{and}\quad \sigma\subs{max}(t) &= \sigma(t,\; t\subs{fail} = (a\subs{min}-a\subs{prev})/\kappa,\; a\subs{min}).
\end{align}


We introduce the additional simplification that variations in penalty costs at different $a\subs{next}$ for sub-trajectories within $t \in [t\subs{now}, t\subs{now} + \hat{\Delta} t]$ are negligible. Their positional difference is (based on the values in Tab.~\ref{tab:referencevalues}) bounded by
\begin{equation}
v_0\,t - \left( v_0\, t + \tfrac{1}{2}\, a\subs{max} \, \hat{\Delta} t \right) \leqslant \tfrac{1}{2} \, a\subs{max} \, \hat{\Delta} t^2 = 4.5\,\unit{cm},
\end{equation}
which, again, is likely far lower than accuracies in environment modeling. In turn, if $\hat{\Delta} t > t\subs{valve}$ (as applies here), we may simplify the statements in \eqref{eq:area-p-b-tau} and \eqref{eq:area-p-c-tau} to
\begin{equation}
{\langle P(..., a\subs{next}, ...)\rangle^{\mathbf B}}' =   \underbrace{\int_{t\subs{now} + \hat{\Delta} t}^{t\subs{hzn}} \hspace{-0.5cm} \dee t
\;\;
\int_{\sigma^{\mathbf{B}}(t, t\subs{now})}^{\sigma^{\mathbf{B}}(t, t\subs{valve})} \hneg\hneg \dee s 
 \;  W(t,s)
\hspace{0.5pt}\left.\frac{\partial \tau\subs{fail}^{\mathbf{B}}}{\partial s}\right|_{t,s}}_{\mathbf{B}_2}
\end{equation}

\begin{align}
{\langle P\rangle^{\mathbf{C}}_{\boldsymbol\varepsilon}}' &=
\underbrace
{
\int_{t\subs{now}+\hat{\Delta}t}^{t\subs{plan}}\hneg\dee t\quad
\int_{\sigma^{\mathbf{C}}(t, t\subs{valve})}^{\sigma^{\mathbf{B}}(t, t-\varepsilon_4)}\hneg\dee s\;
W(t,s)\left.\frac{\partial \tau^{\mathbf{C}}}{\partial s}\right|_{t,s}
}_{\mathbf C_1}\notag\\
&+
\underbrace
{
\int_{t\subs{plan}+\varepsilon_6}^{t\subs{hzn}}\hneg\,\dee t\quad
\int_{\sigma^{\mathbf{C}}(t, t\subs{valve})}^{\sigma^{\mathbf{B}}(t, t\subs{plan})}\hneg\dee s\;
W(t,s)\left.\frac{\partial \tau^{\mathbf{C}}}{\partial s}\right|_{t,s}
}_{\mathbf C_2}
\end{align}
which eliminates the term over $\mathbf{B}_1$, as well as the limits for $\varepsilon_1$, $\varepsilon_2$, $\varepsilon_3$ and $\varepsilon_5$.

\subsection{Area under $\mathbf{B}$}

We precompute the antiderivative \emph{relative to the initial trajectory} $\sigma(t, t\subs{fail}=t\subs{now})$ (which decelerates with $a\subs{prev}$ and is hence invariant to $a\subs{next}$), by using
\begin{align}
W^{\mathbf{B}}(\hat{t},\hat{s}) &= W(\hat{t},\hat{s}) \left.\frac{\partial \tau\subs{fail}^{\mathbf{B}}}{\partial s}\right|_{\hat{t},\hat{s}}\\
\text{as}\quad \mathcal I^{\mathbf{B}}(\hat{t},\hat{s}) &= - W^{\mathbf{B}}(\hat{t},\sigma(\hat{t}, t\subs{now})) + \sum_{\mathclap{ \set{ s \in \hat S(t) \,|\, s \leqslant \hat{s} } }} W^{\mathbf{B}}(\hat{t},s). 
\end{align}
Due to this, we then can evaluate a given $a\subs{next}$ via
\begin{equation}
P^{\mathbf{B}}(\hat{t}, a\subs{next}) = |\; \mathcal I^{\mathbf{B}}(\hat{t},\sigma(\hat{t}, \hat{t}\subs{valve}, a\subs{next}))\; | .
\label{eq:p-B}
\end{equation}

%

\subsection{Area under $\mathbf{C}$}

For $\partial \tau\subs{fail}^{\mathbf{C}}/\partial s$ as in \eqref{eq:dtau-c-ds} we distinguish between its domain $\mathbf{C}_\blacktriangleright$ (where it depends on $a\subs{next}$) and $\mathbf{C}_\blacksquare$ (where it does not). We hence define the following arrays which are both \emph{invariant} with $a\subs{next}$:
\begin{align}
W^{\mathbf{C}_\blacktriangleright}(\hat{t},\hat{s}) &= \sqrt{-a\subs{next}}\,W(\hat{t}, \hat{s}) \left.\frac{\partial \tau\subs{fail}^{\mathbf{C}}}{\partial s}\right|_{\hat{t},\hat{s}} \!\!\!\!\!= \frac{W(\hat{t}, \hat{s})}{\sqrt{2\,v_0\,\hat{t}-2\,\hat{s}}}
\\
\text{and } W^{\mathbf{C}_\blacksquare}(\hat{t},\hat{s}) &= W(\hat{t},\hat{s}),
\end{align}
which can be accumulated to give the antiderivatives
\begin{equation}
\mathcal I^{\mathbf{C}_\blacktriangleright}(\hat{t},\hat{s}) = \!\sum_{\mathclap{ \set{ s \in \hat S(t) \,|\, s \leqslant \hat{s} } }} W^{\mathbf{C}_\blacktriangleright}(\hat{t},s)
\;\,\text{and}\;\,
\mathcal I^{\mathbf{C}_\blacksquare}(\hat{t},\hat{s}) = \!\sum_{\mathclap{ \set{ s \in \hat S(t) \,|\, s \leqslant \hat{s} } }} W^{\mathbf{C}_\blacksquare}(\hat{t},s).
\end{equation}
These can then be used to evaluate a given $a\subs{next}$: For the case of $\sigma(t, t\subs{fail}, a\subs{next})$ lying entirely within $\mathbf{C}_\blacktriangleright$ for all $t\subs{fail} \in [t\subs{valve}, t\subs{plan}]$, we have
\begin{equation}
\begin{aligned}
P^{\mathbf{C}_\blacktriangleright}(\hat{t}, a\subs{next}) = \frac{1}{\sqrt{-a\subs{next}}}
\,
\bigg(\;\;\;&\mathcal I^{\mathbf{C}_\blacktriangleright}(\hat{t}, \sigma(\hat{t}, \hat{t}\subs{plan}, a\subs{next}))\\[-7pt]
- &\mathcal I^{\mathbf{C}_\blacktriangleright}(\hat{t}, \sigma(\hat{t}, \hat{t}\subs{valve}, a\subs{next})) \;\;\;\bigg).
\end{aligned}
\label{eq:p-C-play}
\end{equation}
For the case of $\sigma(t, t\subs{fail}, a\subs{next})$ lying entirely within $\mathbf{C}_\blacksquare$ for all $t\subs{fail} \in [t\subs{valve}, t\subs{plan}]$, we have
\begin{equation}
\begin{aligned}
P^{\mathbf{C}_\blacksquare}(\hat{t}, a\subs{next}) = \frac{1}{v_0}
\,
\bigg(\;\;\;&\mathcal I^{\mathbf{C}_\blacksquare}(\hat{t}, \sigma(\hat{t}, \hat{t}\subs{plan}, a\subs{next}))\\[-7pt]
- &\mathcal I^{\mathbf{C}_\blacksquare}(\hat{t}, \sigma(\hat{t}, \hat{t}\subs{valve}, a\subs{next})) \;\;\;\bigg)\,.
\end{aligned}
\label{eq:p-C-stop}
\end{equation}
For cases which transition between $\mathbf{C}_\blacktriangleright$ and $\mathbf{C}_\blacksquare$ by crossing $[s_\blacksquare, t_\blacksquare]$ (cf. Fig.~\ref{fig:trajseparation}), we evaluate each side separately.

\section{PRACTICAL RESULTS}\label{sec:practical-results}

The algorithm was evaluated on different platforms using different parameters. For the risk predictions in $W(t,s)$, both purely synthetic noise fields were used, as well as data from simulated traffic scenarios.


\begin{figure}%
\vspace{4pt}

\begin{subfigure}{\columnwidth}%
\centering
\begin{overpic}{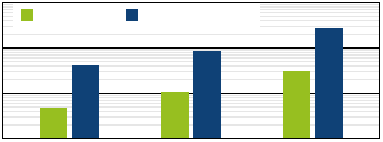}%
\put(11,31.5){\footnotesize proposed}
\put(38.5,31.5){\footnotesize direct}
\put(-13,9.5){\parbox{2cm}{\raggedleft\footnotesize 4.8\,ms}}
\put(18,21){\parbox{2cm}{\raggedright\footnotesize 41.2\,ms}}
\put(18,14){\parbox{2cm}{\raggedleft\footnotesize 10.9\,ms}}
\put(50,25){\parbox{2cm}{\raggedright\footnotesize 87.7\,ms}}
\put(50.00, 20){\parbox{2cm}{\raggedleft\footnotesize 30.6\,ms}}
\put(82.00, 31){\parbox{2cm}{\raggedright\footnotesize 272.1\,ms}}
\put(3,-5){\parbox{2cm}{\centering\footnotesize i7-2600}}
\put(34.5, -5){\parbox{2cm}{\centering\footnotesize i5-470UM}}
\put(66.00, -5){\parbox{2cm}{\centering\footnotesize Raspberry Pi 3}}
\put(-17,0){\parbox{1cm}{\raggedleft\footnotesize $0\,\unit{ms}$}}
\put(-17.00, 11){\parbox{1cm}{\raggedleft\footnotesize $10\,\unit{ms}$}}
\put(-17.00, 23){\parbox{1cm}{\raggedleft\footnotesize $100\,\unit{ms}$}}
\put(-17.00, 35){\parbox{1cm}{\raggedleft\footnotesize $1000\,\unit{ms}$}}
\end{overpic}\\[7pt]
\caption{Worst-case execution times on different systems on a logarithmic scale. The proposed algorithm reduces the computation time consistently to about 11\,\% of that of the direct computation.}%
\label{fig:laufzeit-system}%
\end{subfigure}\\[10pt]

\begin{subfigure}{\columnwidth}%
\centering
\begin{overpic}{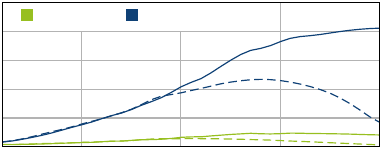}%
\put(11,33.5){\footnotesize proposed}
\put(38.5,33.5){\footnotesize direct}
\put(-7,-5){\parbox{1cm}{\centering\footnotesize $2\,\unit{m/s}$}}
\put(14,-5){\parbox{1cm}{\centering\footnotesize $10\,\unit{m/s}$}}
\put(40.00, -5.00){\parbox{1cm}{\centering\footnotesize $20\,\unit{m/s}$}}
\put(66.00, -5.00){\parbox{1cm}{\centering\footnotesize $30\,\unit{m/s}$}}
\put(92,-5){\parbox{1cm}{\centering\footnotesize $40\,\unit{m/s}$}}
\put(-17,0){\parbox{1cm}{\raggedleft\footnotesize $0\,\unit{ms}$}}
\put(-17.00, 7.60){\parbox{1cm}{\raggedleft\footnotesize $10\,\unit{ms}$}}
\put(-17.00, 15.20){\parbox{1cm}{\raggedleft\footnotesize $20\,\unit{ms}$}}
\put(-17.00, 22.80){\parbox{1cm}{\raggedleft\footnotesize $30\,\unit{ms}$}}
\put(-17.00, 30.40){\parbox{1cm}{\raggedleft\footnotesize $40\,\unit{ms}$}}
\put(-17,38){\parbox{1cm}{\raggedleft\footnotesize $50\,\unit{ms}$}}
\end{overpic}\\[10pt]
\caption{Execution times on an i7-2600 over different $v_0$, for a maximum allowed braking distance of 100\,m (dashed) and 200\,m (solid). A shorter maximum distance reduces computation times at high $v_0$ because gentler decelerations (causing longer braking distances) can be ruled out \emph{a-priori}.}%
\label{fig:laufzeit-wall}%
\end{subfigure}

\caption{Comparison of execution times between the proposed algorithm (green) and the direct computation by various relevant factors.}%
\label{fig:laufzeit}%
\end{figure}

\begin{figure}%

\begin{subfigure}{\columnwidth}
\centering
\begin{overpic}[
]{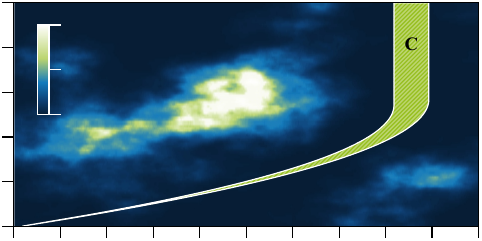}%
\put(-3,-3){\parbox{1cm}{\centering\footnotesize $0\,\unit{m}$}}
\put(7,-3){\parbox{1cm}{\centering\footnotesize $10\,\unit{m}$}}
\put(17,-3){\parbox{1cm}{\centering\footnotesize $20\,\unit{m}$}}
\put(27,-3){\parbox{1cm}{\centering\footnotesize $30\,\unit{m}$}}
\put(36,-3){\parbox{1cm}{\centering\footnotesize $40\,\unit{m}$}}
\put(46,-3){\parbox{1cm}{\centering\footnotesize $50\,\unit{m}$}}

\put(46,-7){\parbox{1cm}{\centering\footnotesize $s$}}

\put(56,-3){\parbox{1cm}{\centering\footnotesize $60\,\unit{m}$}}
\put(64.5,-3){\parbox{1cm}{\centering\footnotesize $70\,\unit{m}$}}
\put(74,-3){\parbox{1cm}{\centering\footnotesize $80\,\unit{m}$}}
\put(83,-3){\parbox{1cm}{\centering\footnotesize $90\,\unit{m}$}}
\put(92,-3){\parbox{1cm}{\centering\footnotesize $100\,\unit{m}$}}
\put(-13,2){\parbox{1cm}{\raggedleft\footnotesize $0\,\unit{s}$}}
\put(-13,11){\parbox{1cm}{\raggedleft\footnotesize $2\,\unit{s}$}}
\put(-13,21){\parbox{1cm}{\raggedleft\footnotesize $4\,\unit{s}$}}

\put(-17,25.5){\parbox{1cm}{\raggedleft\footnotesize $t$}}

\put(-13,30){\parbox{1cm}{\raggedleft\footnotesize $6\,\unit{s}$}}
\put(-13,39){\parbox{1cm}{\raggedleft\footnotesize $8\,\unit{s}$}}
\put(-13,49){\parbox{1cm}{\raggedleft\footnotesize $10\,\unit{s}$}}

\put(14,25.5){\parbox{1cm}{\raggedright\footnotesize\color{white} $0.0$}}
\put(14,34.5){\parbox{1cm}{\raggedright\footnotesize\color{white} $0.5$}}
\put(14,43.5){\parbox{1cm}{\raggedright\footnotesize\color{white} $1.0$}}
\put(23,43.5){\parbox{1cm}{\raggedright\footnotesize\color{white} $W(t,s)$}}

\end{overpic}\\[13pt]
\caption{Result at $v_0 = 30\,\unit{\frac{m}{s}}$: To avoid high values of $W(t,s)$, the optimal choice is to maintain the valve setting at $a\subs{prev} = a\subs{next} = -5.5\,\unit{m/s^2}$. Since the valve does not transition, \textbf{B} is empty; the width of the enclosed area only results from the vehicle motion within $[t\subs{now},t\subs{plan}]$, corresponding to set \textbf{C}.}%
\end{subfigure}\\[15pt]

\begin{subfigure}{\columnwidth}
\centering
\begin{overpic}[
]{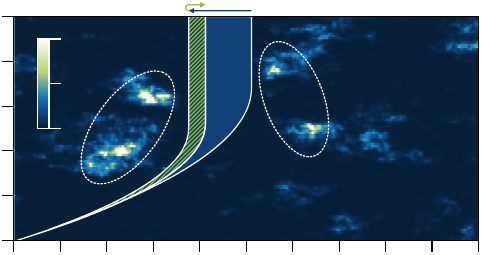}%
\put(-3,-3){\parbox{1cm}{\centering\footnotesize $0\,\unit{m}$}}
\put(7,-3){\parbox{1cm}{\centering\footnotesize $10\,\unit{m}$}}
\put(17,-3){\parbox{1cm}{\centering\footnotesize $20\,\unit{m}$}}
\put(27,-3){\parbox{1cm}{\centering\footnotesize $30\,\unit{m}$}}
\put(36,-3){\parbox{1cm}{\centering\footnotesize $40\,\unit{m}$}}
\put(46,-3){\parbox{1cm}{\centering\footnotesize $50\,\unit{m}$}}

\put(46,-7){\parbox{1cm}{\centering\footnotesize $s$}}
\put(40,52){\parbox{1cm}{\centering\footnotesize $t\subs{fail}$}}

\put(56,-3){\parbox{1cm}{\centering\footnotesize $60\,\unit{m}$}}
\put(64.5,-3){\parbox{1cm}{\centering\footnotesize $70\,\unit{m}$}}
\put(74,-3){\parbox{1cm}{\centering\footnotesize $80\,\unit{m}$}}
\put(83,-3){\parbox{1cm}{\centering\footnotesize $90\,\unit{m}$}}
\put(92,-3){\parbox{1cm}{\centering\footnotesize $100\,\unit{m}$}}
\put(-13,2){\parbox{1cm}{\raggedleft\footnotesize $0\,\unit{s}$}}
\put(-13,11){\parbox{1cm}{\raggedleft\footnotesize $2\,\unit{s}$}}
\put(-13,21){\parbox{1cm}{\raggedleft\footnotesize $4\,\unit{s}$}}

\put(-17,25.5){\parbox{1cm}{\raggedleft\footnotesize $t$}}

\put(-13,30){\parbox{1cm}{\raggedleft\footnotesize $6\,\unit{s}$}}
\put(-13,39){\parbox{1cm}{\raggedleft\footnotesize $8\,\unit{s}$}}
\put(-13,49){\parbox{1cm}{\raggedleft\footnotesize $10\,\unit{s}$}}

\put(14,25.5){\parbox{1cm}{\raggedright\footnotesize\color{white} $0.0$}}
\put(14,34.5){\parbox{1cm}{\raggedright\footnotesize\color{white} $0.5$}}
\put(14,43.5){\parbox{1cm}{\raggedright\footnotesize\color{white} $1.0$}}

\put(23,43.5){\parbox{1cm}{\raggedright\footnotesize\color{white} $W(t,s)$}}

\end{overpic}\\[13pt]
\caption{Result at $v_0 = 15\,\unit{\frac{m}{s}}$: The algorithm decides to transition from $a\subs{prev} = -2.2\,\unit{m/s^2}$ to $a\subs{next} = -3.0,\unit{m/s^2}$ to avoid the high penalties in areas (dashed ellipses). The largest area of $W(t,s)$ is swept by the transition; in contrast to (a), later $t\subs{fail}$ produce shorter stopping distances as the stronger deceleration outweighs the vehicle's motion at $v_0$.}%
\end{subfigure}\\[8pt]

\begin{subfigure}{\columnwidth}
\centering
\begin{overpic}[
]{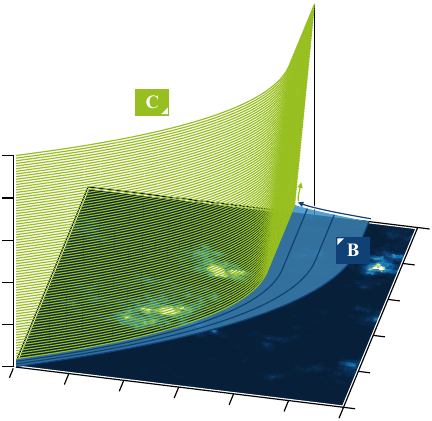}%
\put(-15,11){\parbox{1cm}{\raggedleft\footnotesize $0.00\,\unit{s}$}}
\put(-15,21){\parbox{1cm}{\raggedleft\footnotesize $0.05\,\unit{s}$}}
\put(-15,31){\parbox{1cm}{\raggedleft\footnotesize $0.10\,\unit{s}$}}

\put(-17,36){\parbox{1cm}{\raggedleft\footnotesize $t\subs{fail}$}}

\put(-15,41){\parbox{1cm}{\raggedleft\footnotesize $0.15\,\unit{s}$}}
\put(-15,51){\parbox{1cm}{\raggedleft\footnotesize $0.20\,\unit{s}$}}
\put(-15,61){\parbox{1cm}{\raggedleft\footnotesize $0.25\,\unit{s}$}}
%
\put(-5.00, 5.00){\parbox{1cm}{\centering\footnotesize $0\,\unit{m}$}}
\put(7.50, 3.50){\parbox{1cm}{\centering\footnotesize $10\,\unit{m}$}}
\put(20.00, 2.00){\parbox{1cm}{\centering\footnotesize $20\,\unit{m}$}}
\put(32.50, 0.50){\parbox{1cm}{\centering\footnotesize $30\,\unit{m}$}}

\put(30.50, -5){\parbox{1cm}{\centering\footnotesize $s$}}

\put(45.00, -1.00){\parbox{1cm}{\centering\footnotesize $40\,\unit{m}$}}
\put(57.50, -2.50){\parbox{1cm}{\centering\footnotesize $50\,\unit{m}$}}
\put(70.00, -4.00){\parbox{1cm}{\centering\footnotesize $60\,\unit{m}$}}

\put(84.00, 1.00){\parbox{1cm}{\raggedright\footnotesize $0\,\unit{s}$}}

\put(87.40, 9.40){\parbox{1cm}{\raggedright\footnotesize $2\,\unit{s}$}}
\put(90.80, 17.80){\parbox{1cm}{\raggedright\footnotesize $4\,\unit{s}$}}

\put(98, 21){\parbox{1cm}{\raggedright\footnotesize $t$}}

\put(94.20, 26.20){\parbox{1cm}{\raggedright\footnotesize $6\,\unit{s}$}}
\put(97.60, 34.60){\parbox{1cm}{\raggedright\footnotesize $8\,\unit{s}$}}

\put(101.00, 43.00){\parbox{1cm}{\raggedright\footnotesize $10\,\unit{s}$}}

\end{overpic}\\[5pt]
\caption{Result of (a), with $t\subs{fail}$ on a separate axis: The transition area \textbf{B} is wide, but swept only for $0.008\,\unit{s}$ or $3.2\,\%$ of the time, due to the relatively minor change in valve state. It overlaps with the area \textbf{C} of constant motion with $v_0$, such that along $t\subs{fail}$, several stopping distances $s$ are attained twice, once if the valve fails in transition, and once after it has reached $a\subs{next}$.}%
\end{subfigure}
\caption{Examples of optimization results on Brownian noise fields, as used in \cite{fabianma}, illustrating the effects of constant vehicle motion (a) if the valve state is not changed; and of valve transition (b, c) leading to reoccurring stopping distances at increasing $t\subs{fail}$. Results of realistic traffic scenarios can be found in \cite{tuevnotbrems}.}%
\label{fig:noise}%
\end{figure}
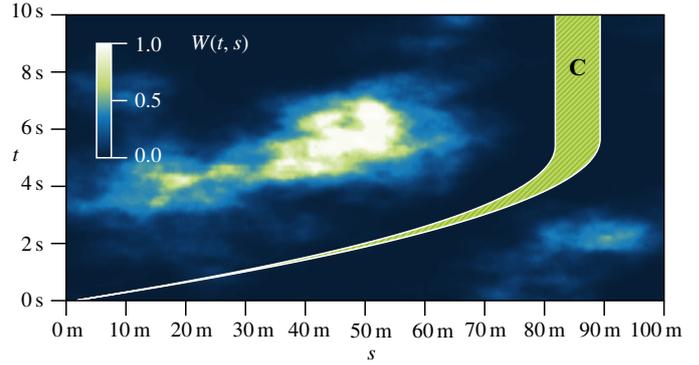
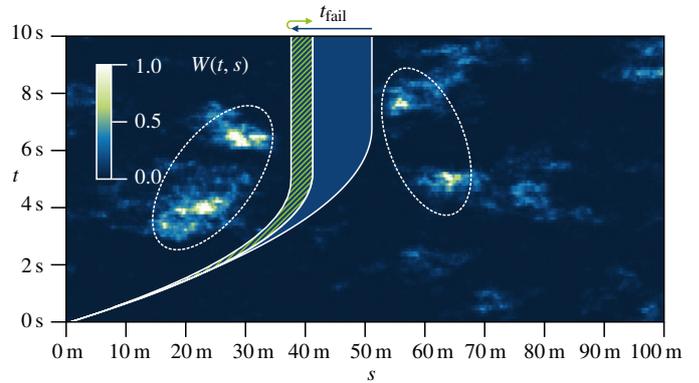
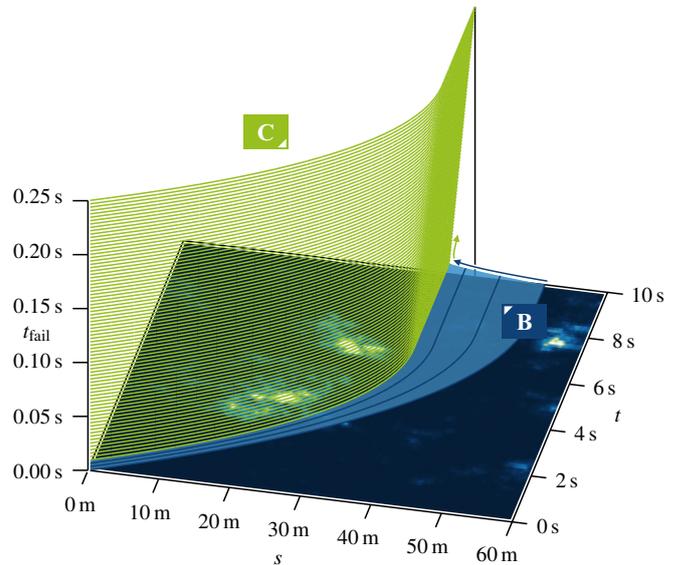

To verify the results and to relate computation speeds, the proposed implementation described here was compared to a \emph{direct} solution of \eqref{eq:planningproblem}: To achieve comparable result quality, $t\subs{fail}$ was discretized into $\hat{T}\subs{fail}$ such that, during a transition, any intermediate arc length $\hat{s} \in \hat S$ is evaluated at least once, as is the case with the proposed algorithm. For each $a\subs{next} \in \hat A$, the \emph{direct} solver computes and minimizes $\langle P \rangle$ by \eqref{eq:expected-value}, and therein computes $P(..., t\subs{fail}, ...)$ by \eqref{eq:penaltyfunction}; the \emph{proposed} solver precomputes $\mathcal I^{\mathbf{B}}$, $\mathcal I^{\mathbf{C_\blacktriangleright}}$,  $\mathcal I^{\mathbf{C_\blacksquare}}$, and then for each $a\subs{next} \in \hat A$, computes and minimizes $P^{\mathbf{B}}+P^{\mathbf{C}_\blacktriangleright}+P^{\mathbf{C}_\blacksquare}$, accumulated over $\hat{t}$, via \eqref{eq:p-B}, \eqref{eq:p-C-play}, \eqref{eq:p-C-stop}.

Experimental results using random Brownian noise fields for $W(t,s)$ (cf. Fig.~\ref{fig:noise}) and simulated traffic scenarios (in \cite{tuevnotbrems}), show that the \emph{numerical results} of both approaches agree within the numerical tolerance, such that both can, in particular, be used equivalently to obtain the optimization result $a^*$. The key goal of the proposed optimization method, however, is to achieve a significant decrease in \emph{computational effort} with respect to the direct solution.

Formally we note that the algorithms have fundamentally different worst-case complexities: If memory for $W(t,s)$ is not considered, the direct algorithm can work with negligible space, opposed to the proposed algorithm that stores several intermediate results of size $\mathcal O(|\hat S|\,|\hat T|)$. However, the computation time is considerably higher to achieve accurate results: The effort of $\mathcal O(|\hat A| \, |\hat T| \, |\hat{T}\subs{fail}|)$ corresponds to $\mathcal O(|\hat A|\, |\hat T| \, |\hat S|)$, if $|T\subs{fail}|$ is chosen to provide accurate results as described above; in contrast, the proposed algorithm provides the accurate global solution at $\mathcal O(|\hat S|\,|\hat T|+|\hat A|\,|\hat T|)$.


In practical scenarios, this corresponds to an approximate average factor of $8$ in computation time between the proposed and the direct solver (or a reduction of about 89\,\%, Fig.~\ref{fig:laufzeit-system}) when tested on an Intel i7-2600 processor (at 3.4 GHz base clock speed and 3.8 GHz turbo clock speed), an Intel i5-470UM processor (1.33 GHz base, 1.86 GHz turbo) and an ARM Cortex A53 (1.2 GHz in a Raspberry Pi 3). Besides the stated complexity parameters, effort also depends on vehicle speed $v_0$: At a given maximum braking distance $s\subs{max}$, gentler decelerations can be ruled out at higher speeds, since their trajectories would exceed $s\subs{max}$. Either solver can considerably reduce effort by truncating the search space this way, as shown in Fig.~\ref{fig:laufzeit-wall}.

%

\section{CONCLUSION AND OUTLOOK}\label{sec:conclusion}


We have presented the problem of planning an optimal decision for a fail-safe emergency stop system, which can adjust a single hydraulic parameter $a\subs{next}$ that governs the braking deceleration in the event of a failure. This predictive approach allows to adapt the deceleration to the environment of the automated vehicle, and yet does not require any E/E components after failure. Optimization of $a\subs{next}$ has to take into account that the exact time of failure is unknown, leading to uncertainty even about the resulting deceleration due to transition times of hydraulic valve, and thereby to a complex planning task; at the same time, planning must be lightweight since it only serves a system of last resort. Based on these considerations, we have posed a suitable problem model, and restated it to enable efficient precomputation. The implementation provides accurate, globally optimal solutions, yet at only about $\nicefrac{1}{8}$ of the computational effort of a direct solver.

\subsection*{Outlook}

So far, the algorithm's performance was evaluated only on a limited set of scenarios; a more exhaustive evaluation with different vehicle models, traffic scenarios and regular planning / prediction systems is required.

The algorithm yields a globally optimal $a^*$ conditioned on the failure within $[t\subs{now}, t\subs{next}]$ the current situation; this can, however, still lead to non-optimal results outside of the given model: Accelerations applied upon $t\subs{fail}$ in the present model only are included as added longitudinal uncertainties, like sensor and road friction uncertainties. An explicit acceleration model is left for future work. Also, a vehicle approaching a railway crossing or intersection would, initially, pick increasing decelerations $a^*$, until a safe stop \emph{before} the crossing cannot be assured; then it would switch to gentler $a^*$, to clear the crossing before stopping. This leads to a systematic, artifactual selection of marginal decelerations, in between which possible failure trajectories necessarily lie on the crossing; it can be resolved by extending the proposed solver to anticipate future planning intervals.

The present algorithm handles the unknown interval of $t\subs{fail}$ by precomputing antiderivatives; a different approach to reduce the computational effort is to modify the underlying system such that the braking pressure can only be released at several discrete (instead of continuous) $t\subs{fail}$; thereby, no contiguous areas must be evaluated but discrete trajectories.

With the increasing use of GPUs for perception, prediction and planning tasks, it may be desirable to locate the emergency maneuver planner there as well. The highly parallel structure of the proposed solution suggests that an even more lightweight (with respect to other tasks) GPU implementation is possible, and should be evaluated.


\bibliographystyle{myIEEEtran}
\bibliography{rootDuerr_DOI}

\end{document}